\documentclass{article}
\usepackage{ijcai24}

\usepackage{times}
\usepackage{soul}
\usepackage{url}
\usepackage[hidelinks]{hyperref}
\usepackage[utf8]{inputenc}
\usepackage[small]{caption}
\usepackage{graphicx}
\usepackage{amsmath}
\usepackage{amsthm}
\usepackage{booktabs}
\usepackage{algorithm}
\usepackage{algorithmic}
\usepackage[switch]{lineno}
\usepackage{amsfonts}
\usepackage{xcolor}

\usepackage{multirow}
\usepackage{float}
\usepackage{subfig}

\newtheorem{theorem}{Theorem}

\usepackage{paralist}

\newtheorem{assumption}{Assumption}

\title{Personalized Federated Instruction Tuning via Neural Architecture Search
}

\author{
Pengyu Zhang$^1$
\and
Yingbo Zhou$^1$\and
Ming Hu$^{2}$\and
Junxian Feng$^1$\and
Jiawen Weng$^1$\And
Mingsong Chen$^1$
\\
\affiliations
$^1$Shanghai Key Lab of Trustworthy Computing, East China Normal University\\
$^2$Nanyang Technological University\\
}

\begin{document}

\maketitle

\begin{abstract}
Federated Instruction Tuning (FIT) has shown the ability to achieve collaborative model instruction tuning among massive data owners without sharing private data. 
However, it still faces two key challenges, i.e., data and resource heterogeneity.
Due to the varying data distribution and preferences among data owners, FIT cannot adapt to the personalized data of individual owners.
Moreover, clients with superior computational abilities are constrained since they need to maintain the same fine-tuning architecture as the weaker clients.
To address these issues, we propose a novel \textbf{Per}sonalized \textbf{F}ederated \textbf{I}nstruction \textbf{T}uning (\textbf{PerFIT}) framework based on architecture search.
Specifically, PerFIT allows each client to search for a personalized architecture by expanding the trainable parameter space of the global model followed by pruning the parameters to the original state.
This procedure allows personalized instruction fine-tuning within expanded parameter spaces, concurrently preserving the same number of trainable parameters. 
Furthermore, to release the abilities of heterogeneous computational resources and enhance the performance of personalization on local data, we exploit personalized parameter-wise aggregation. 
The evaluation with multiple LLMs non-IID scenarios demonstrates that compared to the state-of-the-art FIT methods, our approach can achieve up to a $23\%$ decrease in perplexity.

\end{abstract}

\section{Introduction}
The emergent abilities of Large Language Models (LLMs) \cite{touvron2023llama} have presented the powerful capability of solving various language-related tasks, including reasoning, text generation, and question-answering. To obtain better-aligned LLMs that can precisely follow the instructions of humans, Instruction Tuning (IT) \cite{wei2022finetuned,wang2022self} has been proposed and demonstrated essential effectiveness in enhancing the generalizability of the foundation LLMs to downstream tasks. Compared to the conventional Fine Tuning (FT) methods, IT incorporates the vanilla text with specific instructions paired with corresponding answers, thereby unlocking the existing abilities of LLMs during the tuning process. 

Though IT is superior to traditional FT, the success of IT greatly relies on the variety, quality, and quantity of the training data. Moreover, the increasing concerns about data privacy \cite{gupta2022recovering} and the expensive expenses of data collecting and cleaning jointly impede the obtaining of large amounts of valuable data. 
Worse still, the heterogeneity of private data fails to reflect the meaningful statistical property of the domain, resulting in the implantation of inevitable bias during IT. 
To overcome the aforementioned issues,  Federated Instruction Tuning (FIT) \cite{zhang2023towards} was introduced as the first exploration of the instruction-based optimization framework in Federated Learning (FL).
The framework enables the effective utilization of computational resources of local devices,  leveraging their private instruction-following data. 
Furthermore, parameter-efficient fine-tuning methods \cite{hu2021lora,lester2021power} have been seamlessly integrated into the FIT framework, enhancing the facilitation of lightweight local tuning processes.


Although the privacy-guaranteed FIT framework can alleviate the data heterogeneity and allow collaboratively training, the preference of local data is not taken into consideration. Moreover, the existing FIT method ignores resource heterogeneity since every client has to share the same structure of fine-tuning modules, potentially causing the waste of resources on clients with larger capabilities. 
To address the challenges of handling local data and resource heterogeneity \cite{ilhan2023scalefl}, we propose an adaptive personalized federated instruction tuning method to enable local clients to fully use their data and resources. Our method is motivated by the intrinsic connection between data heterogeneity and architecture heterogeneity, thereby allowing each client to search for a personal IT architecture. 
Specifically, we adopt the efficient foresight pruning method based on the Taylor expansion of the loss to simplify the expensive Neural Architecture Search (NAS) \cite{mellor2021neural} process. Benefiting from the data-guided pruning, each client owns a personal sparse structure of the IT modules that fit the personalized local data. Furthermore, we propose a personalized aggregation mechanism that achieves parameter-wise aggregation across clients to enhance the information interactions.
Our contributions are summarized as follows:
\begin{itemize}
    \item We propose a novel \textbf{Per}sonalized \textbf{F}ederated \textbf{I}nstruction \textbf{T}uning (\textbf{PerFIT}) method based on neural architecture search, where each client can obtain a tailored fine-tuning architecture according to resource capabilities.

    \item We propose a personalized aggregation strategy for the fine-tuned modules to promote information interaction across local clients with various architectures.
    
    \item We implement our \textbf{PerFIT} framework on well-known LLMs for comprehensive experiments in both resource heterogeneity and homogeneity scenarios, which adequately show the effectiveness of our method.
\end{itemize}

\section{Related Work}

\noindent\textbf{Instruction Tuning of Large Language Models.}
Existing LLMs have demonstrated substantial performance in deriving
task-relevant answers by simply decorating the vanilla input with instructions. However, the fine-tuning process is still a promising option to achieve better results when confronting unexplored tasks \cite{peng2023instruction}. To preserve the advantages of instruction data and fine-tuning, instruction tuning was proposed as an essential approach to optimize the performance of LLMs. This method improves the efficacy of LLMs in handling diverse and complex tasks by fine-tuning them with human instructions and aligning them with real-world tasks \cite{xu2023wizardlm}. 
The benefits include bridging the gap between pertaining objectives and human instructions, enhancing predictability over model behaviors, and refining the resemblance to human-like capabilities and output patterns. Research in this area focuses on two ways to generate instructions: i) prompts manually created by humans \cite{wen2023hard} and ii) instruction-following data auto-generated by machines \cite{wang2022self}.
Expensive as the first method is, the quality of instruction data manufactured with human effects is elevated due to the precise human annotation.
The latter utilizes a self-instruct method based on open-sourced LLMs to auto-generate instruction data. Specifically, a powerful LLM is deployed to generate massive task-specific instruction data, which is subsequently leveraged to boost the alignment ability of another trainable LLM.
However, due to the high value of collecting instruction data for various tasks, the owners of specific data are unlikely willing to share it with other competitors. 
Therefore, the data cross-silo scenarios still exist. The FIT framework proposed by \cite{zhang2023towards} provides a lightweight solution to overcome the challenge brought by decentralized data, but the personalization aspects of local clients including data and resource heterogeneity are not taken into consideration. Therefore, we propose a flexible personalized FIT method, aiming to address both challenges simultaneously.

\noindent\textbf{Personalized Federated Learning.}
Personalized Federated Learning (PFL) focuses on training a client-specific model to achieve satisfying performance for each client instead of a global model to accommodate all client data uniformly.
Specifically, the personalization of clients includes two major aspects: i) data heterogeneity \cite{mendieta2022data} and ii) resource heterogeneity \cite{imteaj2021resource}. The first indicates the differences in local data distributions and the second shows the distinctions in terms of computation abilities, communication overhead, etc.
To address the data heterogeneity challenges, existing methods including \cite{t2020personalized} introduced regularization terms to guide the local objectives.
To tackle the challenge of resource heterogeneity, \cite{shamsian2021hyperperson} proposed to distinguish personalized models from a global model through a hypernetwork.
\cite{yuan2020fednas} derives Federated Neural Network Search (FL-NAS) to obtain personalized architectures based on both data and resource heterogeneity.
Effective as the aforementioned methods are, 
most of them only concentrate on one aspect of personalization. Worse still, none of them are tailored for PFL on LLMs. 
Except for the problem of data heterogeneity, local parameter-efficient fine-tuning on LLMs poses another challenge: the performance of Parameter-Efficient Fine-Tuning (PEFT) on LLMs is related to the specific fine-tuning architecture \cite{lawton-etal-2023-neural}.
Therefore, we propose to utilize the concepts from NAS to connect data heterogeneity to architecture heterogeneity, and further unlock the capability of FIT within various local architectures.

\section{Preliminaries}
\subsection{Personalized Federated Learning}
The goal of PFL is to train personalized models for each client collaboratively.  Considering $n$ clients with private Non-IID dataset denoted as $\mathcal{D}_n=\{(\mathbf{x}_{n,j}, y_{n,j})\}_{j=1}^{N_n}$, we seek to solve the problem below:
\begin{align}
\tiny{
\nonumber\mathop{\arg\min}\limits_{\mathbf{\Theta}}\frac{1}{n}\sum\limits_{i=1}^{n}\mathcal{L}_i(\mathbf{\theta}_i) 
\;,\mathcal{L}_i(\mathbf{\theta}_i) = \frac{1}{N_n}\sum\limits_{j}^{N_n}\ell_i(\mathbf{x}_{n,j}, y_{n,j}; \theta_i).
}
\end{align}
Here, $\theta_i$ represents the trainable parameters of the $i^{th}$ client, $\ell_i$ is the loss function for the $i^{th}$ client, $\mathcal{L}_i(\mathbf{\theta}_i)$ denotes the average loss across the local data. $\mathbf{\Theta} = \{\mathbf{\theta}_i\}_{i=1}^{n}$ represents the set of trainable parameters of personal models.

\begin{figure*}[h]
\centering  
\includegraphics[width=1.0\linewidth]{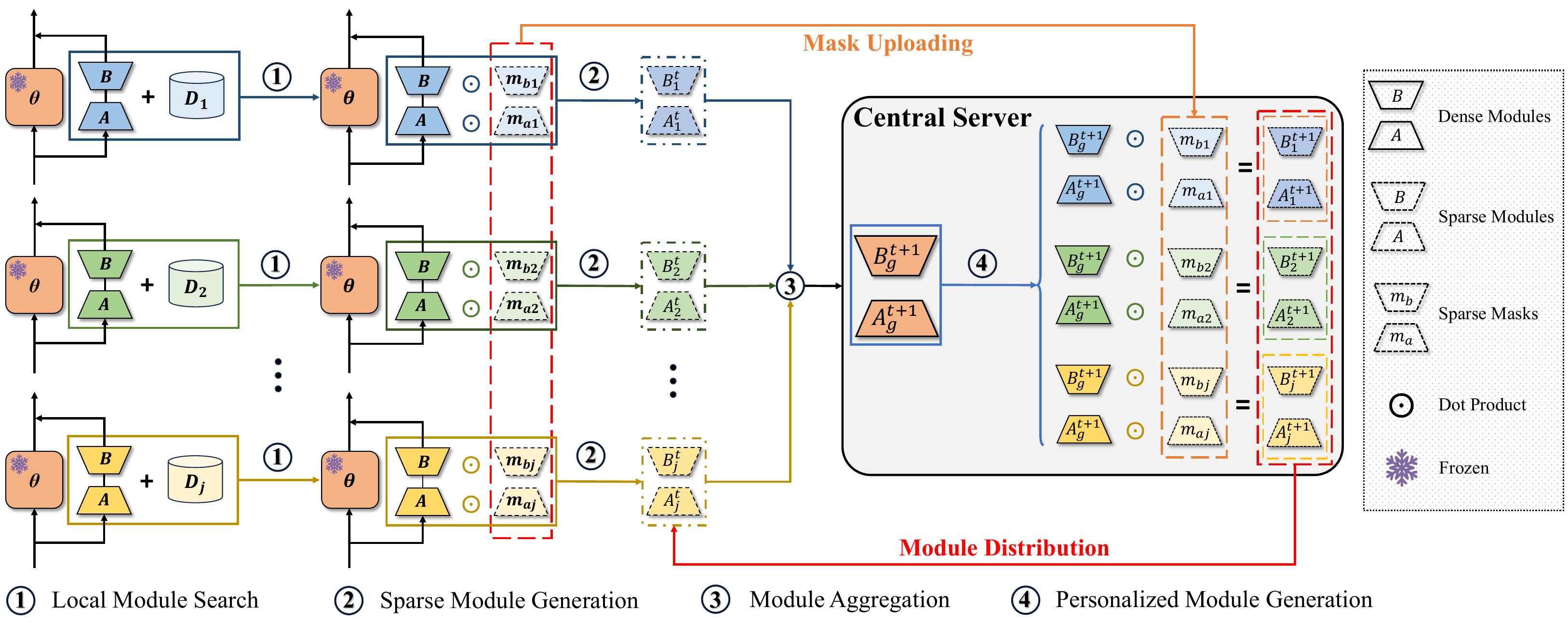} 
\caption{Workflow of our personalized federated instruction tuning approach.}
\label{workflow}
 \vspace{-0.1in}
\end{figure*}

\subsection{Neural Architecture Search (NAS)}
Given a loss function $\ell_i$ and the model parameters $\theta_i(\mathcal{A})$ based on an architecture $\mathcal{A}_i$, we formulate the architecture search as the following optimization problem:
\begin{align}
\label{nas_budget}
\scriptsize
\hspace{-3mm}\mathop{\arg\min}_{\mathcal{A}_i}\ell_{i}(\mathbf{\theta}_i(\mathcal{A}_i); \mathcal{D}_i)
\;\mathop{s.t.}\;
\mathop{R_i(\mathcal{A}_i)}\leq \mathop{B_i}, \; \mathop{i = 1, 2...n}.
\end{align}
Here, $R_i$ and $B_i$ represent the resource consumption and the budget limitation of the $i^{th}$ client. The budget of the $i^{th}$ client can be energy consumption, computational cost, bandwidth requirement, etc., or a combination of these. In this paper, we focus on the number of trainable parameters.
The goal of the NAS is to find a personal training architecture for every client based on the local heterogeneous data $\mathcal{D}_i$.

\subsection{Low-Rank Adapter (LoRA)}
Given the significant constraints on computational resources and communication bandwidth for local clients, we focus on the LoRA method to formulate FIT architectures. 
LoRA achieves the update of fine-tuning by constraining the update of model parameters to maintain a low intrinsic rank. For a pre-trained LLM parameterized by $\mathbf{\theta}_{init}\in \mathbb{R}^{d\times k}$, LoRA utilizes a low-rank decomposition $\mathbf{AB}$ to represent the update $\Delta\mathbf{\theta}$ where $\mathbf{A}\in \mathbb{R}^{d\times r}$, $\mathbf{B}\in \mathbb{R}^{r\times k}$ and the rank $r\ll\mathop{min}(d,k)$. The pre-trained parameter $\mathbf{\theta}$ remains fixed during the fine-tuning while $A$ and $B$ are optimized. The update of $\mathbf{\theta}_{init}$ is formed as
\begin{align}
\nonumber \mathbf{\theta}_{new}\mathbf{x} = \mathbf{\theta}_{init}\mathbf{x} + \Delta\mathbf{\theta x} = \mathbf{\theta}_{init}\mathbf{x}+\mathbf{ABx},
\end{align}
where $\mathbf{\theta}_{new}\in \mathbb{R}^{d\times k}$ denotes the new weight which is re-parameterized after completing the fine-tuning. Note that for mainstream decoder-only LLMs, $d$ equals $k$.

\section{Methodology}
\subsection{Overview of PerFIT}
Figure \ref{workflow} shows the workflow of our method. 
It consists of the following four major steps:
\begin{compactitem}
    \item \textbf{Step 1 (Local Module Search):} Local clients search for their personalized sparse masks. Then, the personalized sparse masks are transmitted to the server.
    \item \textbf{Step 2 (Sparse Module Generation and Local Fine-tuning):} Local clients generate personalized LoRA modules and conduct local fine-tuning.
    \item \textbf{Step 3 (Module Aggregation):} Local clients transmit the sparse fine-tuned LoRA modules to the server. Then the server re-formulates the collected sparse LoRA modules in a dense format and conducts the standard weighted average to obtain global LoRA modules. 
    \item \textbf{Step 4 (Personalized Module Generation and Distribution):} The server generates personalized LoRA modules and distributes them to clients to initialize a new round of local fine-tuning based on the global module and personalized sparse masks.
\end{compactitem}
The backbone of the LLM is frozen during both searching and federated training processes. 
\textbf{Step 1} and \textbf{Step 2} are conducted locally. \textbf{Step 3} and \textbf{Step 4} are conducted in a conventional federated learning manner.
Alg.~\ref{overall_alg} shows the details of the overall workflow, where the ``"Federated Tuning" includes \textbf{Step 2,3} and \textbf{4}. 


\subsection{Implementation Details}
\noindent\textbf{Local Architecture Search through Iterative Pruning.}
For the $i^{th}$ client, we seek to collaboratively search for the personalized architecture $\mathcal{A}_i$ that performs the best on the local dataset $\mathcal{D}_i$. Following Eq.~\ref{nas_budget}, the objective is defined as
\begin{align}
\nonumber&\mathcal{A}_i = 
\mathop{\arg\min}\limits_{\mathcal{A}}\mathcal{L}_i(\mathbf{\theta}_i(\mathcal{A}), \mathcal{D}_i)
\\
&\nonumber\mathop{s.t.}\;
\mathop{R_i(\mathcal{A}_i)}\leq \mathop{B_i}, \mathcal{A}_i\neq\mathcal{A}_j \;\mathop{for} i\neq j,
\end{align}
where $\mathcal{L}_i(\cdot) = \sum_{i=1}^n
p_i\mathcal{L}_i(\cdot)$ and $p_i = |N_n|/\sum_{i=1}^n|N_n|$.
Given the budget of the number of trainable parameters $\mathop{B_i}$, our goal is to find the LoRA architecture $\mathcal{A}_i$ which can achieve the best fine-tuning performance on local data $\mathcal{D}_i$. Due to the heavy burden of traditional NAS on LLMs, we perform the NAS on the LoRA module through foresight iterative pruning. 
Since pruning refers to the process from dense to sparse structure, we first replace the original LoRA module $\mathbf{A}\in \mathbb{R}^{d\times r}$ and 
$\mathbf{B}\in \mathbb{R}^{r\times d}$ with dense $\mathbf{A}_{de}\in \mathbb{R}^{d\times r/(1-s)}$ and 
$\mathbf{B}_{de}\in \mathbb{R}^{r/(1-s)\times d}$, respectively. Note that $s$ represents the sparsity and $0\leq s<1$. While pruning, we aim to remove the elements that have the least impact on the output of the model and reduce the number of parameters from $(d\times r/(1-s))\mathbf{X}$ to $(d\times r)\mathbf{X}$.
To estimate the importance of every element $\theta_i^j$ in $\mathbf{A}_{d}$ and $\mathbf{B}_{d}$, we formulate the change of the loss as 
\begin{align}\label{snip_first}
\vspace{-0.2in}
\textit{I}_{\Delta\theta_i^j} 
\approx \Big|\frac{\partial \ell_i(\Delta\theta_i^j;\mathcal{D}_i)}{\partial \Delta\theta_i^j} \Delta\theta_i^j\Big|,
\end{align}
where $\Delta\theta_i$ is represented by $\mathbf{A}_{de}^i \mathbf{B}_{de}^i$. Eq.~\ref{snip_first} shows the first-order estimation. Similarly, we can derive the parameter-wise second-order estimation as
\begin{align}\label{snip_second}
\textit{I}_{\Delta\theta_i^j} 
\approx \Big|
\theta_i^j H_{jj} \theta_i^j\Big|.
\end{align}
$H$ represents the Hessian matrix and can be approximated by the Fisher information matrix to save the computation burden. In practice, we can use Eq.~\ref{snip_first} or Eq.~\ref{snip_second} or the mixed metric which is defined by
\begin{align}\label{snip_mixed}
\textit{I}_{\Delta\theta_i^j} 
\approx \Big|
\frac{\partial \ell_i(\Delta\theta_i^j;\mathcal{D}_i)}{\partial \Delta\theta_i^j} \Delta\theta_i^j 
-\frac{1}{2} 
\theta_i^j H_{jj} \theta_i^j\Big|.
\end{align}
Since $\mathcal{D}_i$ is the fine-tuning data that has never been used for the pre-training, the two terms in Eq.~\ref{snip_first} and Eq.~\ref{snip_second} are not equal to zero, which shows that the proposed importance score is an ideal measurement of the importance of the architecture of the LoRA modules.
Once we obtain the importance scores, we preserve the parameters that align with the top $100(1-s)\%$  importance scores since these parameters contribute the most to the gradient updates. To avoid the potential layer collapse caused by over-confidence of one-shot pruning, we utilize an exponential decay schedule to complete the pruning within multiple epochs. 
The overall process is described
in Alg.~\ref{alg}.

\begin{algorithm}[h]
\small
    \caption{Neural Architecture Search for LoRA modules} \label{alg}
\textbf{Input}: 
1) $\Delta\theta_0$, dense LoRA module; 
2) $T_p$, \# of pruning epochs; 
3) $m$, \# of total clients;  
4) $s$, sparsity; 
\begin{algorithmic}[1] 
\FOR{$i=1,\dots,m$ in parallel}\label{line:trainStart}
    \FOR{$t=1,\dots,T_p$}
        \STATE Compute $\textit{I}_{\Delta\theta_i}$ based on Eq.~\ref{snip_first} or Eq.~\ref{snip_second} or Eq.~\ref{snip_mixed};
        \STATE Get threshold $\tau$ as $(1-(1-s)^{\frac{t}{T_p}})$ percentile of $\textit{I}_{\Delta\theta_i}$;
        \STATE $\mathbf{m}^i$ as $\mathbf{m}^i\leftarrow\mathbf{m}^i\odot(\textit{I}_{\Delta\theta_i}<\tau)$;
    \ENDFOR
\ENDFOR
\STATE \textbf{Return} Sparse LoRA modules parameterized by $\Delta\theta_i\odot\mathbf{m}^i$
\end{algorithmic}
\end{algorithm}

\noindent\textbf{Symmetric Initialization.}
Different from what was proposed in \cite{hu2021lora}, we conduct the pruning-oriented NAS before starting training to avoid introducing expensive bi-level optimization. However, due to the dependency of the importance measurement on the gradient, we need to carefully initialize the LoRA adapter to prevent \textit{Measurement Vanishing}. Formally, \textit{Measurement Vanishing} indicates that the values of importance scores in equal to zero, resulting in a diminished capability of the metric. 
Since the first and second-order terms in all metrics rely on the gradient, we show that the \textit{Measurement Vanishing} happens without proper initialization.
Based on the chain rule, the gradient of the $\mathbf{A}$ matrix in a LoRA module is defined as $\mathbf{g}_{\mathbf{A}} = \frac{\partial\ell}{\partial o}\mathbf{xB}$.
In standard LoRA configurations, the matrix $\mathbf{B}$ is initialized to all-zeros to avoid adding unexpected perturbations to the frozen backbone model. 
Consequently, the gradient $\mathbf{g}_{\mathbf{A}}$ is zero due to the state of $\mathbf{B}_{de}$. This, in turn, maintains the importance scores $\textit{I}_{\mathbf{A}_{de}}$ at zero, resulting in a consistent pruning of the $\mathbf{A}_{de}$ matrix.
Thus, the \textit{Measurement Vanishing} exists and will undermine the effectiveness of the pruning-oriented NAS process if we keep using the vanilla initialization. Accordingly, we follow the widely-used principle to symmetrically initialize $\mathbf{B}$ with the standard Gaussian and conduct the NAS process based on the following configurations:
\begin{align}
    \nonumber\mathbf{A}_{de}\sim\mathcal{N}(0, 1/d), \;
    \mathbf{B}_{de}\sim\mathcal{N}(0, 1/d),
\end{align}
where $\mathcal{N}$ represents the Gaussian distribution.

\begin{algorithm}[h]
\small
    \caption{Adaptive Personalized FIT} \label{overall_alg}
\textbf{Input}: 
1) $\Delta\theta_0$, dense LoRA module; 
2) $T_p$, \# of pruning epochs; 
3) $T_{tr}$, \# of fine-tuning epochs; 
4) $k$; \# of local clients in each round;
5) $m$, \# of total clients;  
6) $e$, \# of local fine-tuning epochs;
7) $g_s$, a group of sparsity; 
\begin{algorithmic}[1] 
\STATE \textbf{Local LoRA Module Search:}
\FOR{$i=1,\dots,m$ in parallel}
    \STATE Implement Alg.~\ref{alg} based on the $i^{th}$ sparsity in $g_s$.
\ENDFOR
\STATE \textbf{Federated Tuning:}

\FOR{$t=1,\dots,T_{tr}$}
    \STATE $C_k$ $\leftarrow$ $\text{Random\ Sample}$ $k$ clients from $m$ clients; \\
    \STATE $G_k$ $\leftarrow$ Number of elements in $C_k$; \\
    \FOR {$ j = 1,\dots, G_k$ in parallel}
        \STATE Conduct $e$ epochs of local fine-tuning.
    \ENDFOR
    \STATE Upload fine-tuned LoRA modules of clients in $C_k$;
    \STATE Conduct adaptive aggregation based on Eq.~\ref{agg};
    \STATE Dispatch personalized aggregated modules to clients in $C_k$.
\ENDFOR
\STATE \textbf{Return} Sparse LoRA modules parameterized by $\Delta\theta_i\odot\mathbf{m}^i$
\end{algorithmic}
\end{algorithm}

\begin{figure}[h]
\centering  
\includegraphics[width=0.8\linewidth]{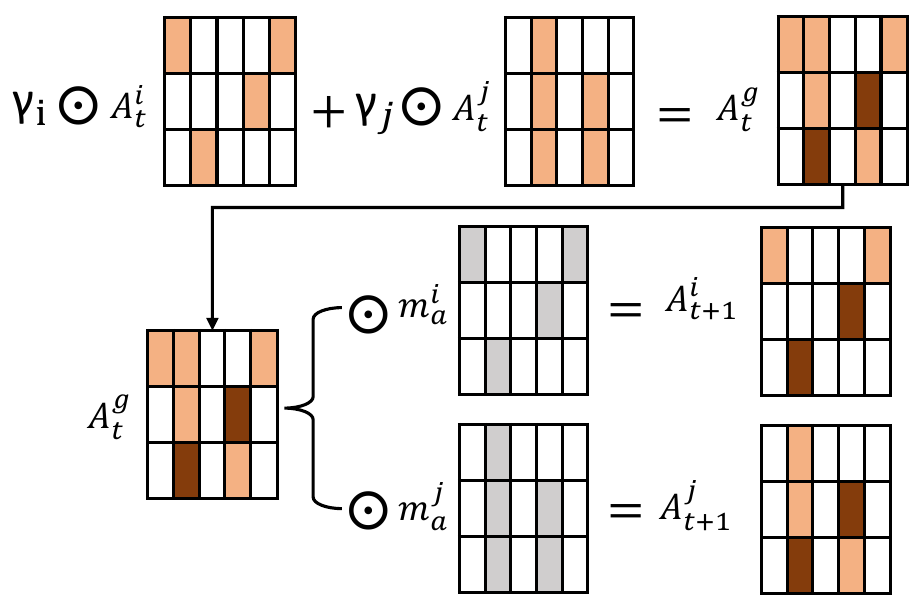} 
\caption{illustration of the personalized aggregation method.}
\label{agg_example}
 \vspace{-0.1in}
\end{figure}

\noindent\textbf{Personalized Aggregation.}
To allow joint optimizations between local trainable parameters in a federated manner, we proposed a personalized aggregation method for the LoRA modules. Formally, we can represent the pruned LoRA modules for the $i^{th}$  client as
\begin{align}\label{mask_form}
    \mathbf{A}_{T=0}^i = \mathbf{A}_{d,T=0}^i \odot \mathbf{m}_a^i, \;
    \mathbf{B}_{T=0}^i = \mathbf{B}_{d,T=0}^i \odot \mathbf{m}_b^i
\end{align}
where $\mathbf{m}_a^i$ and $\mathbf{m}_b^i$ denote the personalized mask matrices given the sparsity $s$. Since the pruning metric defined by Taylor expansion is dependent on the data $\mathcal{D}_i$, the obtained mask matrices vary across clients, i.e., $\mathbf{m}_a^i\neq\mathbf{m}_a^j$ and $\mathbf{m}_b^i\neq\mathbf{m}_b^j$. Intuitively, two personalized masks will not have any overlap if $\mathcal{D}_i$ is strictly heterogeneous to $\mathcal{D}_j$.
Therefore, for a set of local LoRA-$\mathbf{A}$ modules $\{\mathbf{A}^1, \mathbf{A}^2,..., \mathbf{A}^n\}$, we can mark each parameter $\mathbf{A}_{i,j}^k$ in $\mathbf{A}^k$ with two states with respect to the parameter $\mathbf{A}_{i,j}^l$ in $\mathbf{A}^l$: i) ``\textit{exclusive}"; and ii) ``\textit{shared}".
After completing local training, each parameter only performs aggregation with those parameters that are marked as ``\textit{shared}". 
Since we can express the sparse LoRA architecture in the form of the dense matrix defined in Eq.~\ref{mask_form}, the personalized aggregation can be formulated as:
\begin{align}\label{agg}
\scriptstyle
\mathbf{A}^i_{T} = \Biggl(\sum\limits_{j=1}^{n}\gamma_j\mathbf{A}_{T-1}^j\Biggl)\odot\mathbf{m}_a^i
,\;
\mathbf{B}^i_{T} = \Biggl(\sum\limits_{j=1}^{n}\gamma_j\mathbf{B}_{T-1}^j\Biggl)\odot\mathbf{m}_b^i,
\end{align}
where $\gamma_j$ denotes the aggregation coefficient for the $j^{th}$ client.
Figure.~\ref{agg_example} shows an example of aggregation on $\mathbf{A}$ matrices.
In practice, the static sparse masks only need to be transmitted once to the server. The sparse LoRA modules for aggregation can be efficiently transmitted between local clients and the server. Thus, the extra communication overhead is negligible. Based on the adaptive aggregation mechanism between sparse modules, we can further accommodate the fine-tuning process to resource heterogeneity scenarios. Since the main resource bottlenecks for local clients, including memory consumption and FLOPs, are inherently tied to the trainable parameters, we can adapt local module search according to their maximum capability. Such targeted adaptation ensures optimal utilization of resources and empowers the overall performance. Formally, we first conduct Alg.~\ref{alg} based on a group of resource-specific sparsity levels $g_s=\{s_1, s_2,...,s_m\}$. Then, we follow Eq.~\ref{agg} to enable heterogeneous module aggregation.

\subsection{Convergence Analysis}
We present the convergence analysis of the PerFIT. Since our local NAS method is derived from iterative pruning, we demonstrate the proofs from the perspective of sparse federated learning. We make the following assumptions.


\begin{assumption}{(\textbf{Coordinate-wise bounded gradient discrepancy).}}\label{gra_dist}
For any $\Delta\tilde{\mathbf{\theta}}\in \mathbb{R}^{d\times r}$, there exists a constant $C\geq0$ such that $\left\|\nabla \mathcal{L}_i(\Delta\tilde{\mathbf{\theta}})-\frac1m\sum_{j=1}^m\nabla\mathcal{L}_{j}(\Delta\tilde{\mathbf{\theta}})\right\|_\infty\leq C$.
\end{assumption}

\begin{assumption}{(\textbf{Coordinate-wise bounded gradient).}}\label{gra_bound}
The local gradient of each client is bounded by the constant $B$ such that $\|\nabla_{\Delta\tilde{\mathbf{\theta}}}\mathcal{L}_i(\tilde{\boldsymbol{w}})\|_{\infty}\leq B$.
\end{assumption}

\begin{assumption}{(\textbf{Bounded variance).}}\label{var_bound}
The estimated gradient $\boldsymbol{g}_{i,t,\tau}(\Delta\tilde{\mathbf{\theta}}):=\nabla\ell(\Delta\tilde{\mathbf{\theta}})$ at the $\tau^{th}$ local step in the $t^{th}$ round is unbiased such that $\mathbb{E}\left[\left\|\boldsymbol{g}_{i,t,\tau}(\Delta\tilde{\mathbf{\theta}})-\nabla \mathcal{L}_i(\Delta\tilde{\mathbf{\theta}})\right\|^2\right]\leq\sigma^2,\forall i,t,\tau,\Delta\tilde{\mathbf{\theta}}\in\mathbb{R}^{d\times r}$.
\end{assumption}

\begin{assumption}{(\textbf{L-smoothness).}}\label{l_smooth}
The local loss function is L-smoothness such that $\|\nabla \mathcal{L}_i(\Delta\tilde{\mathbf{\theta}}_1)-\nabla \mathcal{L}_i(\Delta\tilde{\mathbf{\theta}}_2)\|\leq L\|\Delta\tilde{\mathbf{\theta}}_1-\Delta\tilde{\mathbf{\theta}}_2\|$ for arbitrary $\Delta\tilde{\mathbf{\theta}}_1$ and $\Delta\tilde{\mathbf{\theta}}_2$ $\in\mathbb{E}^{d\times r}$.
\end{assumption}

\begin{assumption}{(\textbf{(Bounded mask discrepancy).}}\label{mask_dis}
The element-wise discrepancy measured by the hamming distance between any local mask $(dist(\mathbf{m}^i, \mathbf{m}^j))$, between any local search mask and the optimal local mask of it $(dist(\mathbf{m}^i, \mathbf{m}^{i,*}))$, and between any two local optimal masks $(dist(\mathbf{m}^{i,*}, \mathbf{m}^{j,*}))$ are bounded by constants $V$, $Z$ and $U$, respectively.

\end{assumption}



\begin{theorem}{(\textbf{Convergence of PerFIT).}}
Let $N$ and $S$ represent the number of local steps and the number of participants in each round, respectively.
Given the aforementioned assumptions, assume that the learning rate $\eta\leq\frac{1}{16LN}$, the personalized fine-tuning modules $\Delta\tilde{\mathbf{\theta}}_{i,t}$ have the following convergence rate:
\begin{align}
\scriptsize
&\nonumber\frac1{Tm}\sum_{t=0}^{T-1}\sum_{i=1}^{m}\mathbb{E}\left[\left\|\nabla \mathcal{L}_i\left(\Delta\tilde{\mathbf{\theta}}_{i,t}\right)\right\|^2\right]\\
&\leq\frac{3\left(f\left(\Delta\tilde{\mathbf{\theta}}_0\right)-f\left(\Delta\tilde{\mathbf{\theta}}^*\right)\right)}{T\eta N\kappa}+3\rho+\epsilon,
\end{align}
where $\kappa={\frac{1}{2}}-150N^{3}\eta^{3}L^{3}-15N^{2}\eta^{2}L^{2}-5N\eta L$, $\rho=(25N^3\eta^4L^3+\frac{5N^2\eta^3L^2}{2})(\sigma^{2}+18N\Phi)+\frac{4N^{2}\eta^{2}L+N\eta}{2}ZB^{2}
+9N^{2}\eta^{2}L\Phi+\frac{N\eta^{2}L\sigma^{2}}{S},$
$\Phi=(dr/(1-s)-dr)C^{2}+B^{2}(V+Z),$ and $\epsilon=3(dr/(1-s)-dr)C^2+3dr B^2+3UB^2.$
\end{theorem}

\textbf{Assumption~\ref{gra_dist}, \ref{gra_bound}, \ref{var_bound}, and \ref{l_smooth}} follow the commonly used assumptions defined in \cite{huang2022sparseper}. 
Existing work \cite{malladi2023zerofinetuning} has demonstrated that the Hessian of the loss for LLMs shows a small local effective rank, which indicates that the curvature of the loss is constrained along a certain and small number of directions in the parameter space. 
Moreover, the property of effective rank implies that the gradient is more aligned with the directions of higher curvature, which is consequently constrained in certain directions.
Since all local clients share the same frozen backbone model, the curvature differences caused by heterogeneous fine-tuning data are bounded. Note that the NAS metrics defined by Eq.~\ref{snip_first}, Eq.~\ref{snip_second}, and Eq.~\ref{snip_mixed} are based on either gradient or Hessian or both. We assume that the differences in personalized LoRA architectures are bounded as well, which motivates us to make \textbf{Assumption~\ref{mask_dis}}.

\section{Experiments}








\subsection{Experimental Settings}
\textbf{Dataset.}
We conducted our experiments on the Databricks-dolly-15k dataset \cite{DatabricksBlog2023DollyV2}. It is an open-source dataset of instruction-following records generated by Databricks in several behavioral categories, including creative writing, brainstorming, classification, closed QA, generation, information extraction, open QA, and summarization. 
We performed two types of splitting methods to emulate the heterogeneous data distributed to local clients. The first is the \textit{pathological} non-IID setup where each client is randomly assigned 2 classes among 8 total classes. The second non-IID setup follows the \textit{Dichilet} distribution, which is parameterized
by a coefficient $\beta$, denoted as Dir($\beta$). $\beta$ determines the degree of data heterogeneity. The smaller the $\beta$ is, the more heterogeneous the data distributions will be. We set the $\beta$ as 0.5 throughout the experiments.


\noindent\textbf{Models.}
To showcase the effectiveness of our method on various LLMs, we utilized two open-source large language models: Alpaca-7B \cite{alpaca} and Vicuna-7B-v1.5 \cite{vicuna2023}. The two LLMs have been fine-tuned based on the LLaMA \cite{touvron2023llama} to enhance their abilities to understand and respond to human inputs effectively.

\noindent\textbf{Configurations.}
For all experiments, we set the number of total clients as $100$. The backbones of two LLMs are frozen during pruning and local fine-tuning to save the memory. We add LoRA to three attention modules for every layer, i.e., $\textit{Query}$, $\textit{Key}$, and $\textit{Value}$ matrices. For homogeneous resource baselines, we set the rank ($r$) of all LoRA modules as $8$. 
To preserve the same number of trainable parameters as baselines, the sparsity levels for our method are designated as $0.66$, $0.5$, and $0.33$, corresponding to the ranks of $12$, $16$, and $24$.
For heterogeneous scenarios, we categorize the capability of clients into three levels: i) Large; ii) Medium; and iii) Small. Each category owns $1/3$ of the total number of clients. 
We set the rank for clients with the smallest capability as $8$. Therefore, the rank for Medium and Large is set to $12$ and $16$, respectively. For the number of local pruning epochs, we set $10$ to rank $16$ and $5$ for others.
In each round of local fine-tuning, we randomly select $10\%$ of clients. For all experiments, the local batch size is set to $64$. To facilitate training with batched data on a single GPU, we utilize the gradient accumulation with a mini-batch size of $8$. The total training rounds are $30$ for homogeneous scenarios and $50$ otherwise. The local training epoch is 1. We split $80\%$ of local data into training and use the rest to evaluate the performance of personalization.

\subsection{Performance Evaluation}
\noindent\textbf{Performance on Homogeneous Resources.}
Table.~\ref{homo_perp_table} presents the results of the perplexity comparison under homogeneous resources scenarios. The results of the FIT method are obtained by setting the rank to $8$.
Most of the perplexity achieved by implementing our method consistently outperforms the vanilla FIT method. For the pathological none-IID setting,  PerFIT on the Alpaca model with rank $12$, $16$, and $24$ outperforms FIT by $23\%$, $9\%$, and $10\%$, respectively. Under the same non-IID setting, the perplexity results for the Vicuna model with rank $12$, $16$, and $24$ decrease by $3\%$, $3\%$, and $1\%$, respectively. For the Dirichlet ($0.5$) non-IID scenario, our method improves the Alpaca model by $21\%$, $10\%$, and $12\%$, respectively. For the Vicuna model under the Dirichlet setting, our PerFIT method reduces the perplexity by $1\%$, and $1\%$ based on rank $12$ and $16$ settings, respectively. Note that we observe a $5\%$ unexpected increase of perplexity when setting the rank to $24$. We attribute this phenomenon to the different basic abilities of the two foundation models.

\begin{table}[h]
\centering
\begin{tabular}{c|c|c|lc} 
\hline
\multirow{2}{*}{Dis.} & \multirow{2}{*}{Model} & \multirow{2}{*}{Sparsity} & \multicolumn{2}{c}{Methodology} \\ 
\cline{4-5}
 &  &  & \multicolumn{1}{c}{FIT} & PerFIT \\ 
\hline
\multirow{6}{*}{Path.} & \multirow{3}{*}{Alpaca} & $0.33$ & \multirow{3}{*}{$5.15$} & $\textbf{3.93}$$\textbf{(-1.22)}$ \\
 &  & $0.50$ &  & $4.66$$\text{(-0.49)}$ \\
 &  & $0.66$ &  & $4.61$$\text{(-0.54)}$ \\ 
\cline{2-5}
 & \multirow{3}{*}{Vicuna} & $0.33$ & \multirow{3}{*}{$4.22$} & $\textbf{4.09}$$\textbf{(-0.13)}$ \\
 &  & $0.50$ &  & $\textbf{4.09}$$\textbf{(-0.13)}$ \\
 &  & $0.66$ &  & $4.17$$\text{(-0.05)}$ \\ 
\hline
\multirow{6}{*}{\begin{tabular}[c]{@{}c@{}}Dir.\\(0.5)\end{tabular}} & \multirow{3}{*}{Alpaca} & $0.33$ & \multirow{3}{*}{$5.28$} & $\textbf{4.13}$$\textbf{(-1.15)}$ \\
 &  & $0.50$ &  & $4.71$$\text{(-0.57)}$ \\
 &  & $0.66$ &  & $4.61$$\text{(-0.67)}$ \\ 
\cline{2-5}
 & \multirow{3}{*}{Vicuna} & $0.33$ & \multirow{3}{*}{$3.85$} & $3.81$$\text{(-0.04)}$ \\
 &  & $0.50$ &  & $\textbf{3.78}$$\textbf{(-0.07)}$ \\
 &  & $0.66$ &  & $4.05$$\text{(+0.20)}$ \\
\hline
\end{tabular}
\caption{Perplexity comparison. Smaller is better.}
\label{homo_perp_table}
\vspace{-0.1in}
\end{table}

\begin{figure}[h]
 \vspace{-0.1in}
\centering
\subfloat[Path. Alpaca]{\includegraphics[width=.5\linewidth]{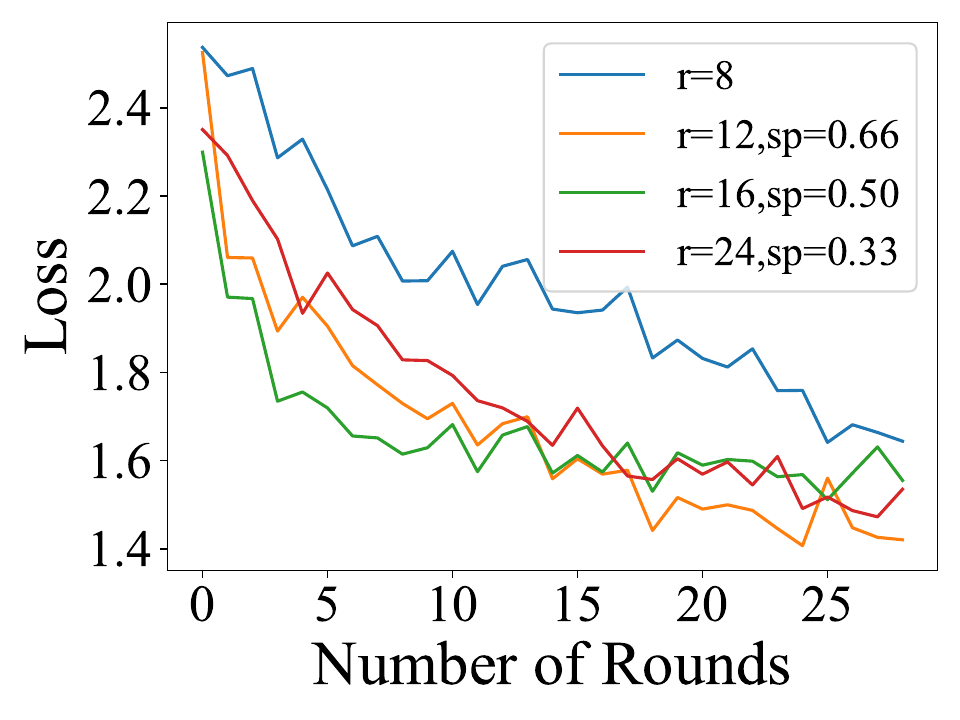}}
\subfloat[Dir. Alpaca]{\includegraphics[width=.5\linewidth]{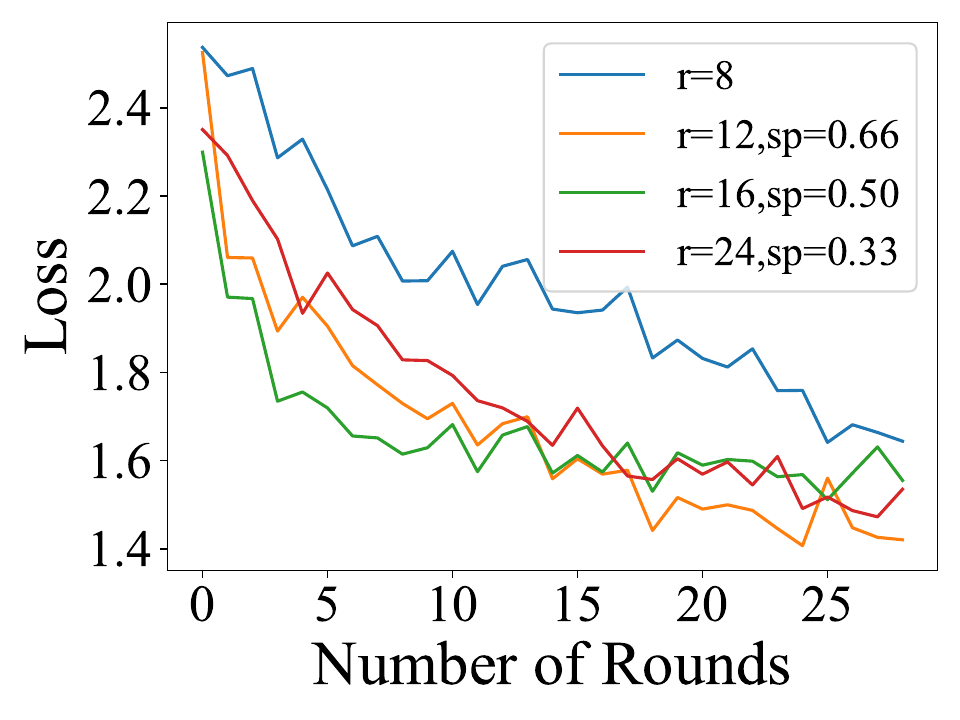}}  \\
\vspace{-0.1in}
\subfloat[Path. Vicuna]{\includegraphics[width=.5\linewidth]{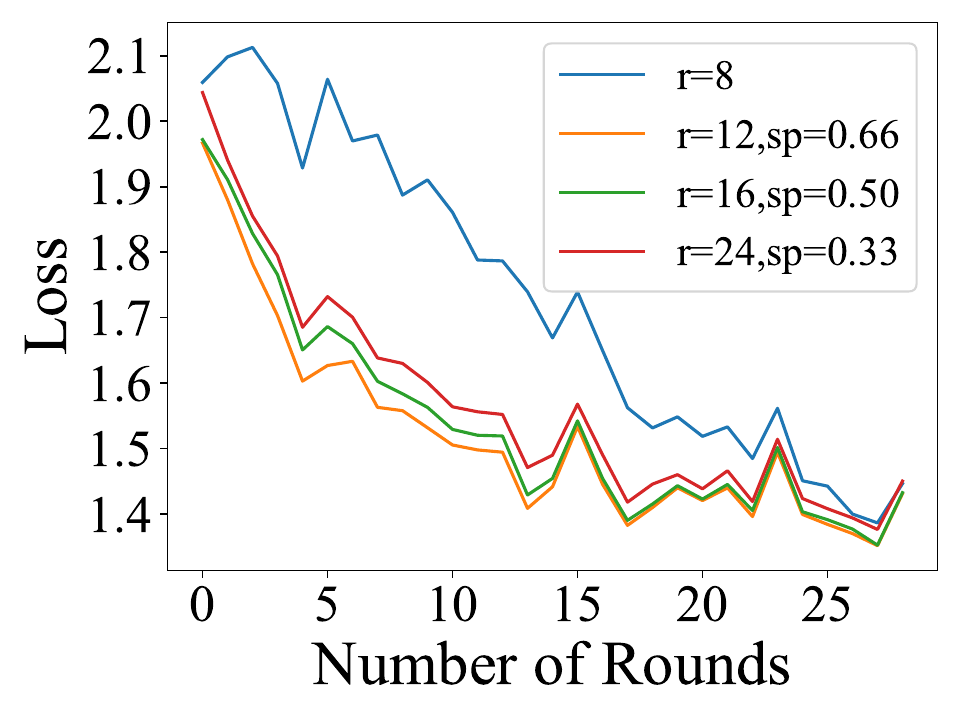}}
\subfloat[Dir. Vicuna]{\includegraphics[width=.5\linewidth]{homo_pdf/path,vicuna.pdf}}
\vspace{-0.1in}
\caption{Loss curves for homogeneous resources. 
}
\label{homo_perp_fig}
\vspace{-0.1in}
\end{figure}

Figure.~\ref{homo_perp_fig} shows the corresponding loss curves. The first row represents the curves when fine-tuning the Alpaca model and the second shows the results of fine-tuning the Vicuna. For the Alpaca model, we consistently observe fast convergence and lower losses with all rank settings. The curve with a rank of $12$ converges to the smallest value of loss on two different non-IID settings.
For the Vicuna model, we find that our PerFIT method invariably enjoys a fast convergence speed at the early stage on all rank settings. The curve with a rank of $12$ exhibits the best overall performances considering both convergence speed and loss value.
In both non-IID settings, we can observe that the initial and final states of Vicuna exhibit smaller loss values and perplexities compared to Alpaca, indicating that Vicuna is more powerful than the Alpaca model.
Therefore, we can conclude that the Vicuna model exhibits a more flat loss landscape based on our observations and their performances on well-known open-sourced benchmarks for generalizations. 
In addition, based on the previously stated assumptions, the personalized performances are more aligned with those in vanilla FIT. This characteristic offers insight into why the loss curves of the Vicuna model tend to converge to a close value. 

\noindent\textbf{Performance on Heterogeneous Resources.}
Table.~\ref{hetero_perp_table} shows the perplexity results of heterogeneous resources. By utilizing the proposed architecture search and personalized aggregation methods, we can observe that the PerFIT method facilitates local fine-tuning within heterogeneous resource scenarios. 
It is worth noting that the Vicuna still behaves better than the Alpaca model on resource heterogeneity scenarios.
Under the pathological non-IID setting, our method shows a $12\%$ decrease in perplexity compared to the FIT when fine-tuning the Alpaca model. For the Vicuna model, we can observe a $3\%$ reduction in perplexity. With Dirichlet configuration, our method improves the perplexity by $2\%$ and $4\%$ on the Alpaca and Vicuna models, respectively.

\begin{table}[h]
\vspace{-0.05in}
\centering
\begin{tabular}{c|c|lc} 
\hline
\multirow{2}{*}{Dis.} & \multirow{2}{*}{Model} & \multicolumn{2}{c}{Methodology} \\ 
\cline{3-4}
 &  & \multicolumn{1}{c}{FIT} & PerFIT \\ 
\hline
\multirow{2}{*}{Path.} & Alpaca & $4.48$ & $\textbf{3.93}$$\textbf{(-0.55)}$ \\ 
\cline{2-4}
 & Vicuna & $3.78$ & $\textbf{3.63}$$\textbf{(-0.15)}$ \\ 
\hline
\multirow{2}{*}{\begin{tabular}[c]{@{}c@{}}Dir.\\(0.5)\end{tabular}} & Alpaca & $4.17$ &  $\textbf{4.05}$$\textbf{(-0.12)}$ \\ 
\cline{2-4}
 & Vicuna & $3.70$ & $\textbf{3.52}$$\textbf{(-0.18)}$ \\
\hline
\end{tabular}
\caption{Perplexity comparison. Smaller is better.}
\label{hetero_perp_table}
\vspace{-0.1in}
\end{table}

\begin{figure}[h]
\vspace{-0.25in}
\centering
\subfloat[Path. Alpaca]{\includegraphics[width=.5\linewidth]{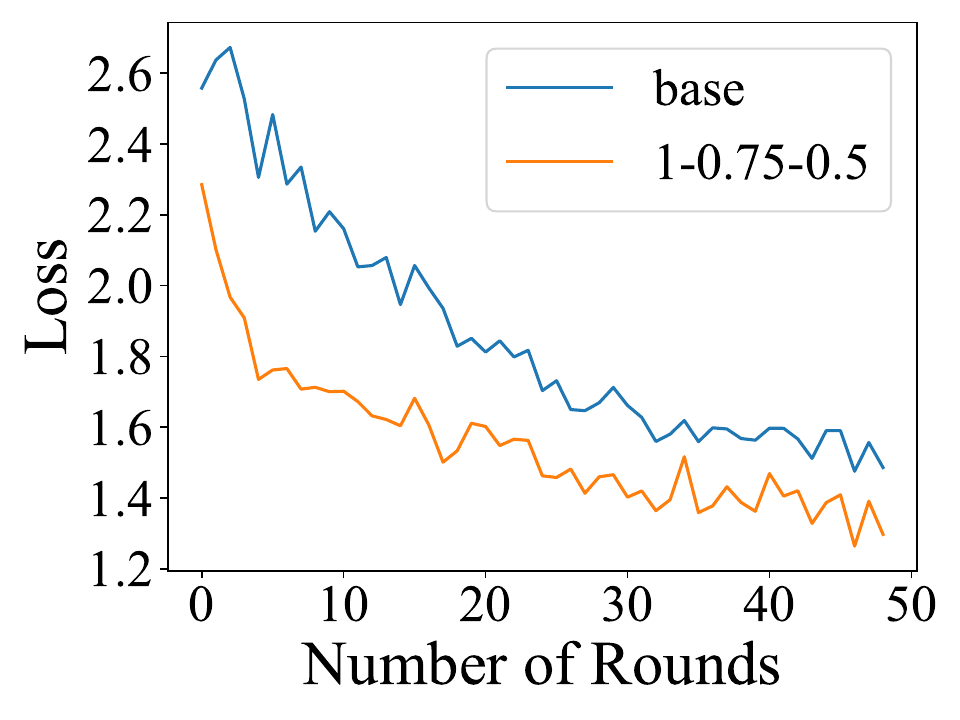}}
\subfloat[Dir. Alpaca]{\includegraphics[width=.5\linewidth]{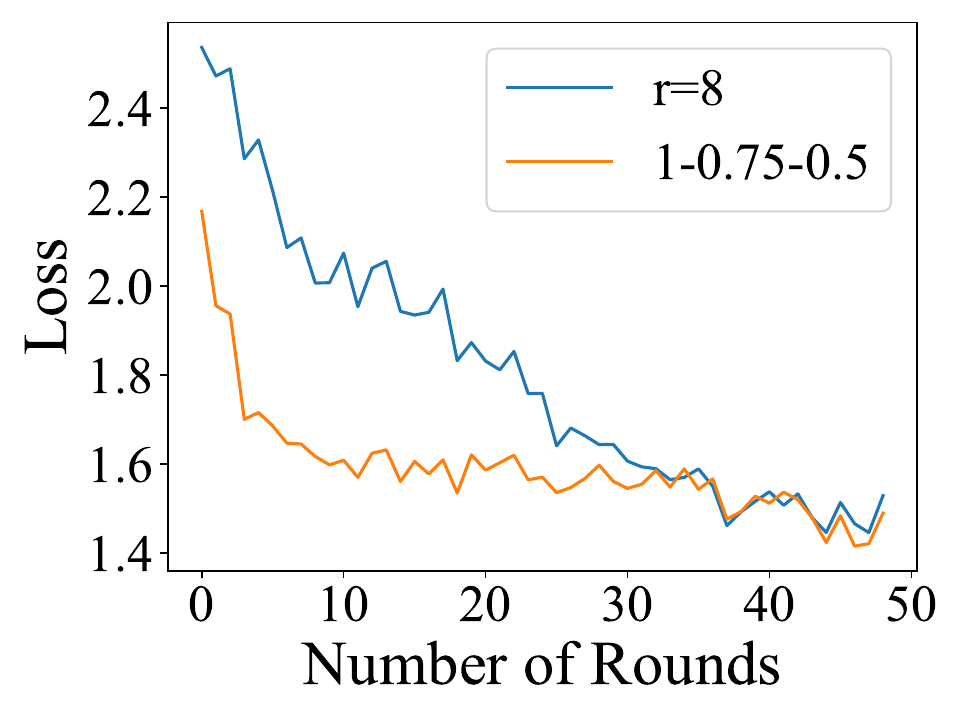}}  \\
\vspace{-0.1in}
\subfloat[Path. Vicuna]{\includegraphics[width=.5\linewidth]{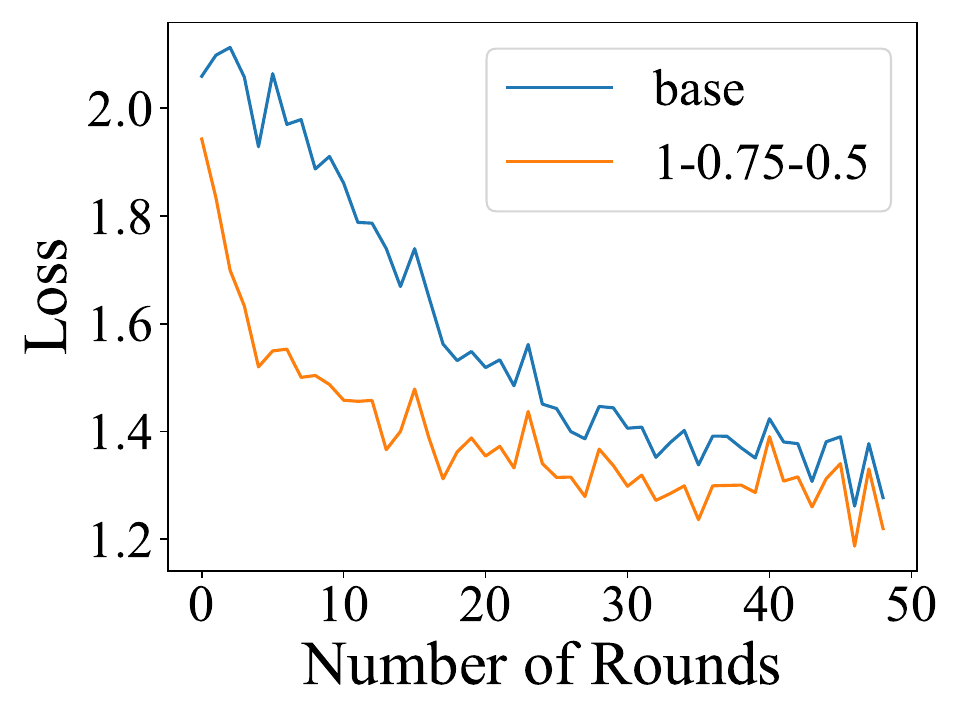}}
\subfloat[Dir. Vicuna]{\includegraphics[width=.5\linewidth]{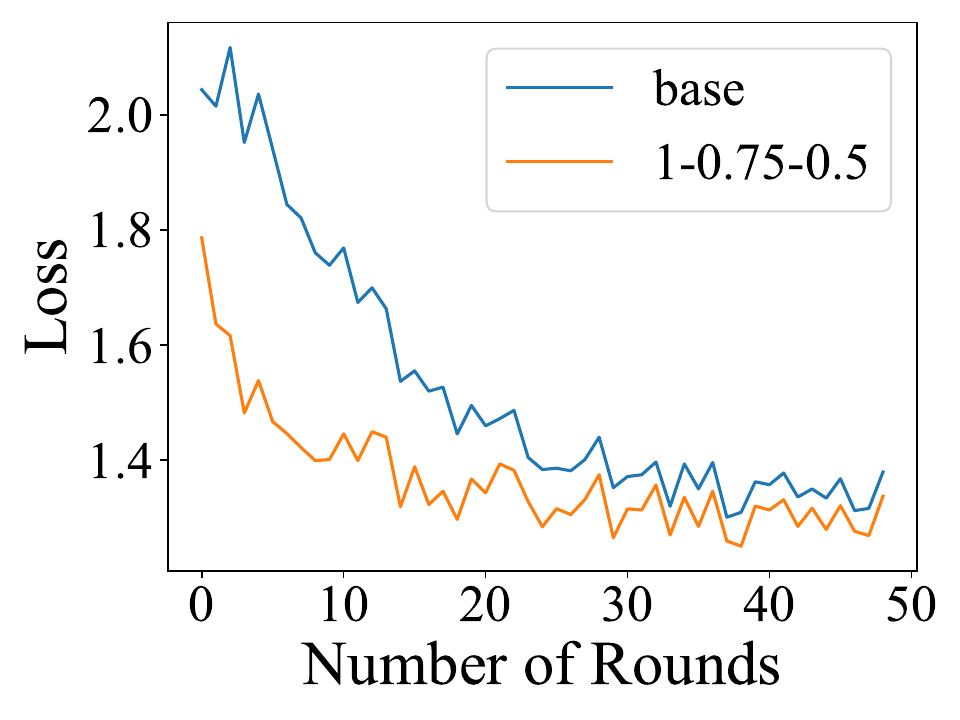}}
\vspace{-0.05in}
\caption{Loss curves for heterogeneous resources.}
\label{hetero_perp_fig}
\vspace{-0.15in}
\end{figure}

Figure.~\ref{hetero_perp_fig} displays the associated loss curves. ``base'' represents the results obtained by setting rank to $8$. ``$1-0.75-0.5$'' represents the performance of our PerFIT method. In this figure, we can observe that our method significantly improves the performance of personalization given the two non-IID scenarios, proving that our method can not only allow collaborative fine-tuning for resource heterogeneous clients but also boost the overall personalization performance.

\begin{figure}[h]
\vspace{-0.1in}
\centering  
\includegraphics[width=0.8\linewidth]{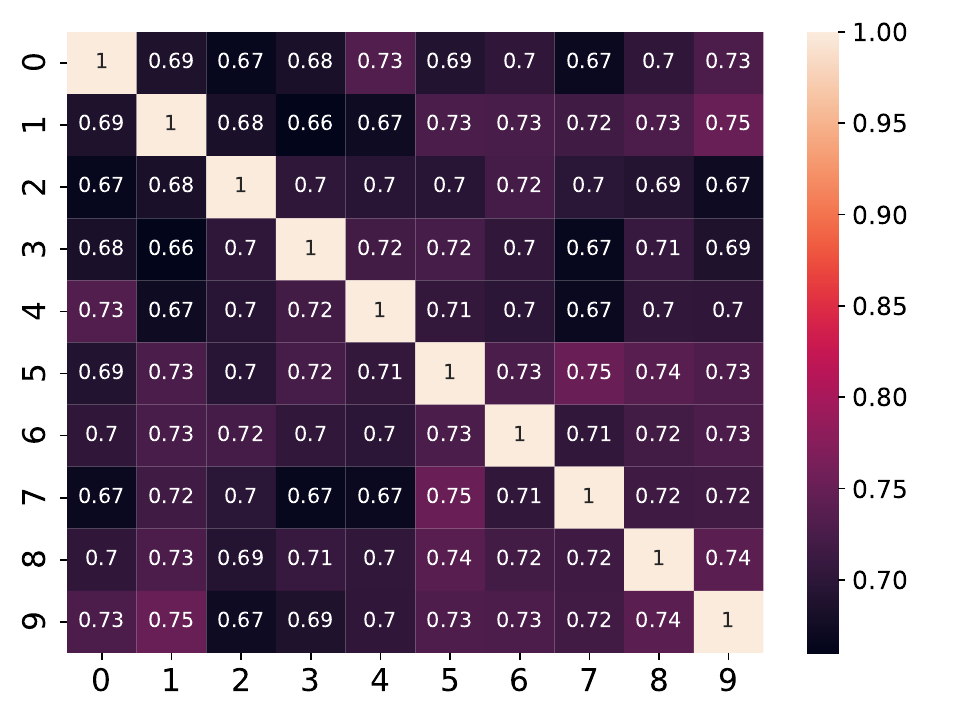} 
\caption{Comparison of different pruning metrics.}
\label{mask_similarity}
 \vspace{-0.1in}
\end{figure}

\noindent\textbf{Mask Similarity Analyses.}
Figure.~\ref{mask_similarity} shows the pair-wise mask similarity between the first LoRA modules of $10$ clients randomly selected. The rank is set to $16$ and the sparsity is set to $0.50$. The labels of the x and y-axes represent the index of the client. The similarity is measured by the hamming distance.
We can observe that clients with heterogeneous data own personalized masks. Furthermore, the degree of any pair-wise similarity is close across clients, which supports and reinforces our assumption of bounded mask discrepancy.

\noindent\textbf{Different Numbers of participants.}
To demonstrate the scalability of our method across various numbers of participants in each round, we conducted extensive experiments by randomly selecting $5\%$ and $20\%$ clients in each round under the Dirichlet non-IID settings.
For the Alpaca model, we can observe that our method 
Similar to the results shown in Figure.~\ref{homo_perp_fig}, we observe that our method implemented on the Alpaca model displays more notable performance improvements. For the Vicuna model, we find that our method converges to the same value as that of FIT but with a remarkable increase in the speed of convergence.

\begin{figure}[h]
\vspace{-0.3in}
\centering
\subfloat[$5$ clients, Alpaca]{\includegraphics[width=.5\linewidth]{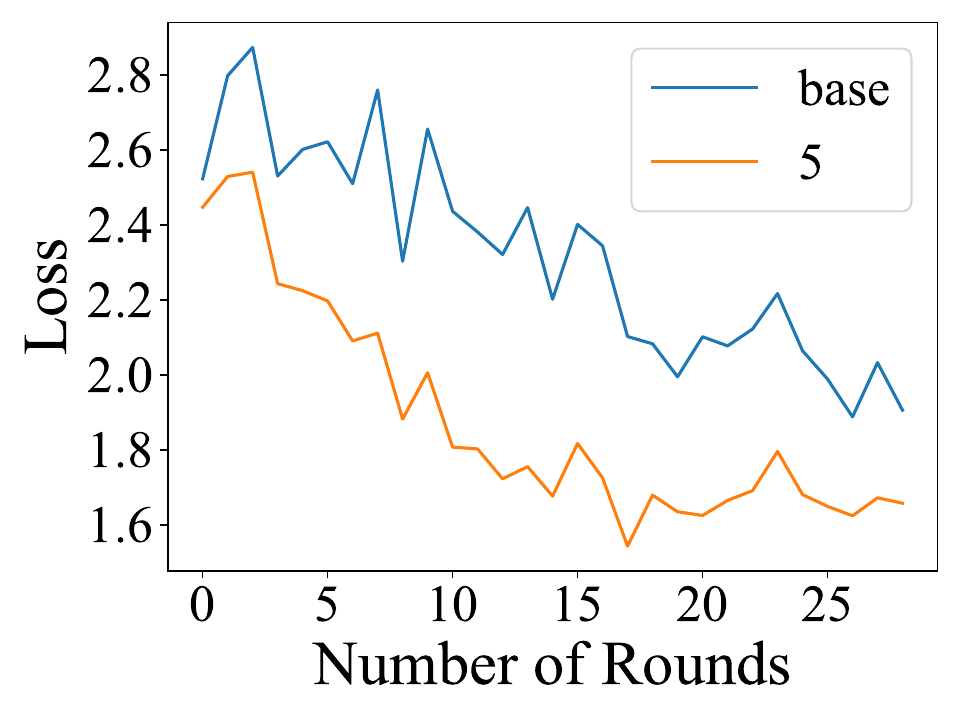}}
\subfloat[$20$ clients, Alpaca]{\includegraphics[width=.5\linewidth]{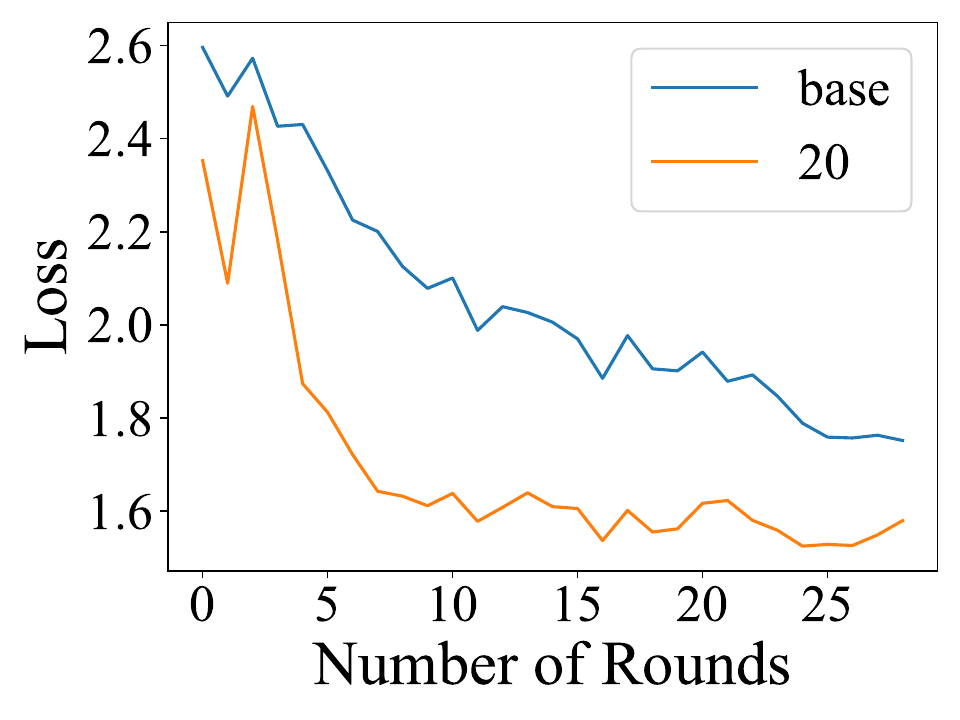}}  \\
\vspace{-0.1in}
\subfloat[$5$ clients, Vicuna]{\includegraphics[width=.5\linewidth]{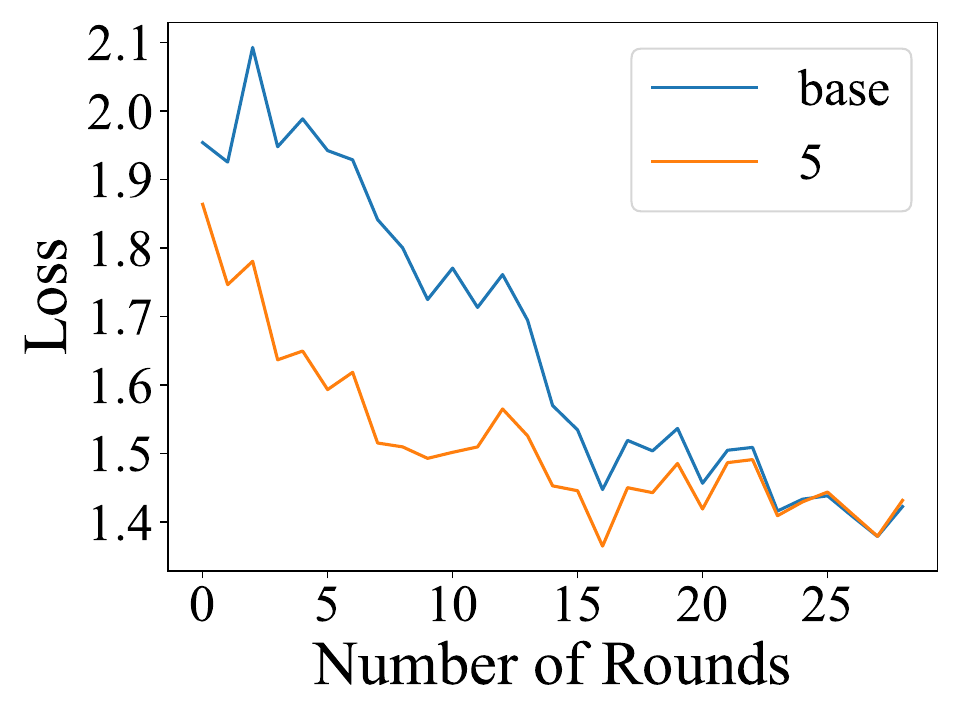}}
\subfloat[$20$ clients, Vicuna]{\includegraphics[width=.5\linewidth]{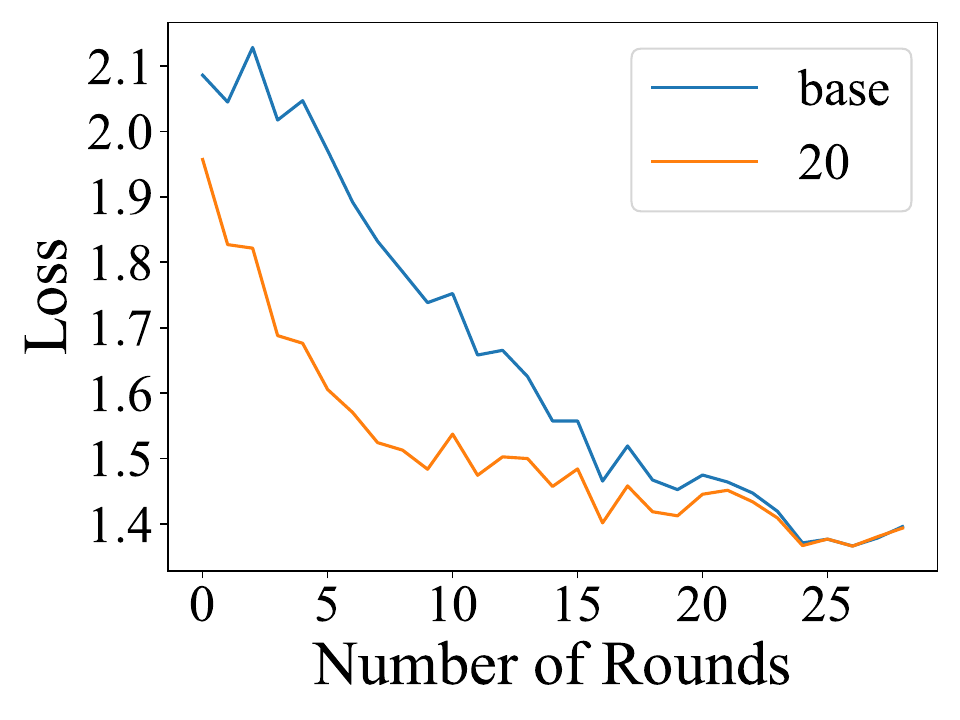}}
\vspace{-0.05in}
\caption{Loss curves of different \# of local participants.}
\label{5_20}
 \vspace{-0.15in}
\end{figure}

\begin{figure}[h]
\vspace{-0.15in}
\centering  
\includegraphics[width=0.58\linewidth]{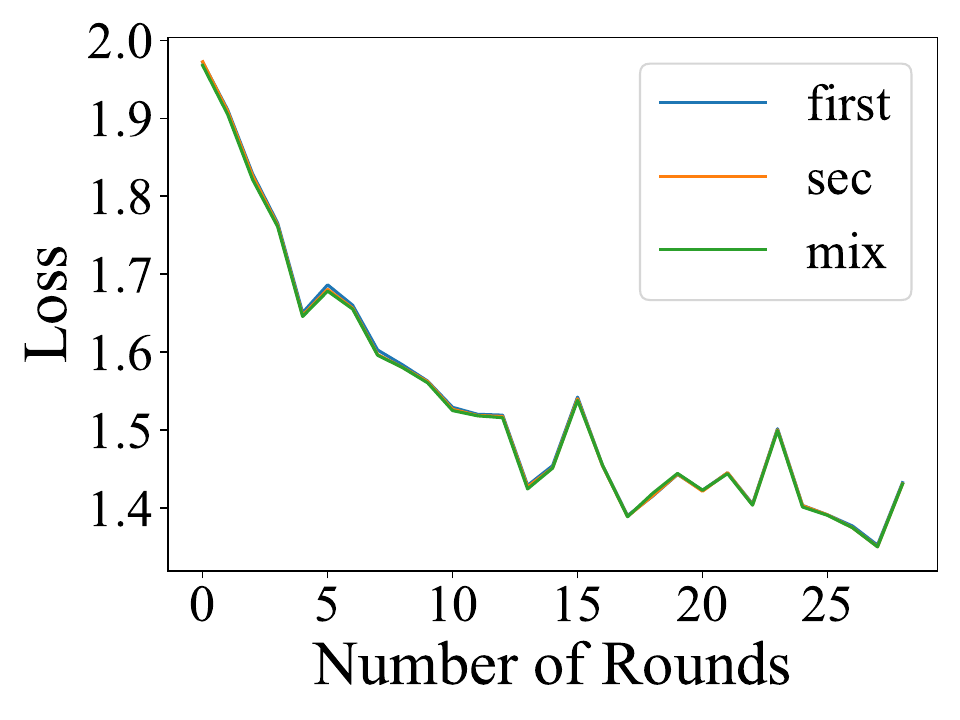} 
 \vspace{-0.1in}
\caption{Comparison of different pruning metrics.}
\label{metric_fig}
 \vspace{-0.2in}
\end{figure}

\noindent\textbf{Impact of Important Score Metric.}
To evaluate the impact of using different metrics to compute the important scores, we further conducted experiments on the Vicuna model under the pathological non-IID settings. The rank is set to $16$ with a sparsity of $50\%$.
The comparisons are shown in Figure.~\ref{metric_fig}. 
The ``first", ``second", and ``mix" curves denote the results obtained based on Eq.~\ref{snip_first}, Eq.~\ref{snip_second}, and Eq.~\ref{snip_mixed}, respectively. We can observe that all metrics exhibit extremely similar behaviors. Since the second-order information requires extra computation overhead, the first-order metric is preferred in practice.

\section{Conclusion}
In this paper, we introduced Personalized Federated Instruction Tuning via Neural Architecture Search. By enabling local clients to search for personalized fine-tuning architectures, we alleviated the challenge arising from data and resource heterogeneity.
We analyzed the convergence property of our method, showing that our method tailored for LLMs exhibits a similar convergence rate to the sparse federated training applied to non-LLMs.
Comprehensive experimental results on representative LLMs under two non-IID scenarios
demonstrated the effectiveness of our proposed method.



\clearpage

\bibliographystyle{named}
\bibliography{ijcai24}

\end{document}